\documentclass[letterpaper, 10 pt, conference]{ieeeconf}  

\IEEEoverridecommandlockouts                              

\usepackage[utf8]{inputenc}
\inputencoding{utf8}
\usepackage{hyperref}
\usepackage{subcaption}
\usepackage{graphics} 
\usepackage{epsfig} 
\usepackage{amsmath} 
\usepackage{amssymb}  
\usepackage{bm}
\usepackage{algorithm}
\usepackage[noend]{algpseudocode}

\title{\LARGE \bf
Towards semi-episodic learning for robot damage recovery
}

\author{Konstantinos Chatzilygeroudis$^{1,2,3}$, Antoine Cully$^{4}$ and Jean-Baptiste Mouret$^{1,2,3}$*
\thanks{\scriptsize*Corresponding author: {\tt\footnotesize jean-baptiste.mouret@inria.fr}}
\thanks{\scriptsize$^1$Inria, Villers-lès-Nancy, F-54600, France}%
\thanks{\scriptsize$^2$CNRS, Loria, UMR 7503, Vandœuvre-lès-Nancy, F-54500, France}%
\thanks{\scriptsize$^3$Université de Lorraine, Loria, UMR 7503, Vandœuvre-lès-Nancy, F-54500, France}%
\thanks{\scriptsize$^4$Personal Robotics Lab, Department of Electrical and Electronic Engineering, Imperial College London, UK}
}

\begin{document}

\maketitle
\thispagestyle{empty}
\pagestyle{empty}

\begin{abstract}

The recently introduced Intelligent Trial and Error algorithm (IT\&E) enables robots to creatively adapt to damage in a matter of minutes by combining an off-line evolutionary algorithm and an on-line learning algorithm based on Bayesian Optimization. We extend the IT\&E algorithm to allow for robots to learn to compensate for damages while executing their task(s). This leads to a semi-episodic learning scheme that increases the robot's life-time autonomy and adaptivity. Preliminary experiments on a toy simulation and a 6-legged robot locomotion task show promising results.
\end{abstract}

\section{Introduction}


Recent research on autonomous systems and robotics has achieved important progress in increasing the autonomy of robots, which makes it possible to operate robots for long periods of time in real-world scenarios. Nevertheless, as robots move from controlled and well-structured environments to more complex \cite{nagatani2013emergency} and more natural ones \cite{Devy19955}, they must be able to react to unforeseen situations; in particular, they have to face the inevitable fact that they will be damaged \cite{carlson_follow-up_2004,carlson_how_2005}.

Current methods for robot damage recovery can be divided into two categories: (1) diagnosis-based approaches \cite{verma_real-time_2004}, and (2) learning methods --- mostly Reinforcement Learning (RL) techniques \cite{ahmadzadeh2014online,erden2008free,cully_robots_2015}. Most of the techniques in the first category require to anticipate the situations that the robot may have to face; an issue can be diagnosed only if the right sensors are present in the right place. These requirements make diagnosis-based techniques difficult to use in complex robotics systems/scenarios --- typically they are only used in the lowest levels of control. Nevertheless, the state-of-the-art RL approaches are also difficult to use for damage recovery because they require many iterations to converge. For example, many RL approaches require tens if not hundreds or thousands of iterations to learn problems with low-dimensional state spaces and fairly benign dynamics, like the mountain car \cite{sutton1998reinforcement}. The data efficiency of RL approaches is a critical aspect that limits their application in real-world robotics scenarios \cite{deisenroth2011pilco}.

A promising approach is the {\it Intelligent Trial and Error algorithm} (IT\&E), a recently introduced algorithm \cite{cully_robots_2015}. The intuition behind IT\&E is that, before the mission, an off-line and computationally expensive evolutionary algorithm can be used to create a behavior-performance map that predicts the performance of thousands of different behaviors. While in mission, this map, guides a fast and on-line search, based on Bayesian Optimization \cite{brochu_tutorial_2010}, to find a compensatory behavior. An important idea is that the behavior-performance map is created using a simulated intact robot, but the algorithm is able to find a working behavior on the damaged real robot because some behaviors from the map perform similarly on the intact and the damaged robot (typically, the behaviors that do not rely on the broken part). The most recent results showed that IT\&E can allow various types of robots (a 6-legged robot and an 8-DOF manipulator) to compensate for many different types of injuries in a matter of minutes \cite{cully_robots_2015,cully2015creative}.

\begin{figure}[!t]
  \centering
  \begin{subfigure}[t]{0.21\textwidth}
    \includegraphics[scale=0.07]{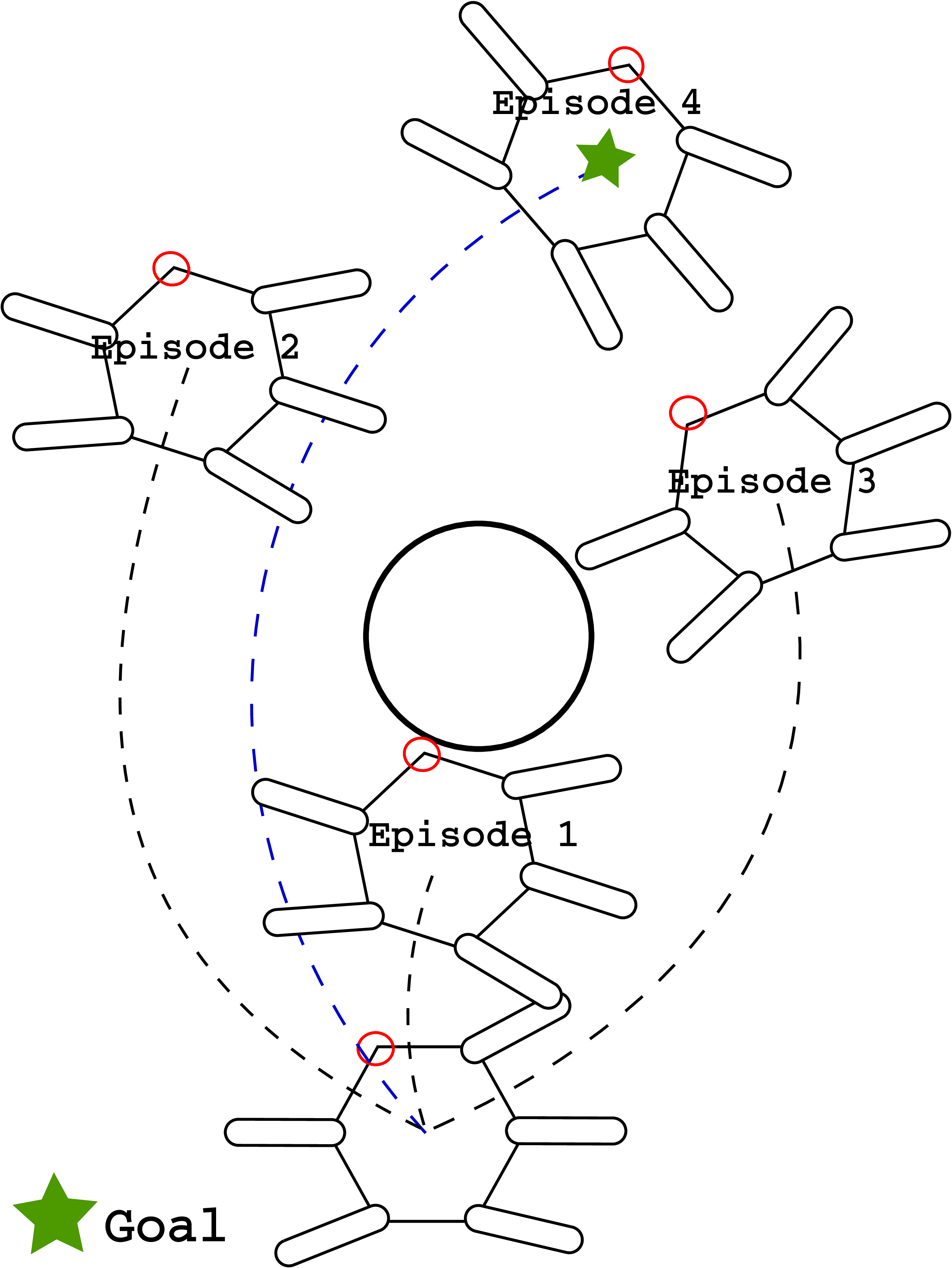}
    \caption{\label{fig:concept_episodic}}
  \end{subfigure} \quad
  \begin{subfigure}[t]{0.2\textwidth}
    \includegraphics[scale=0.07]{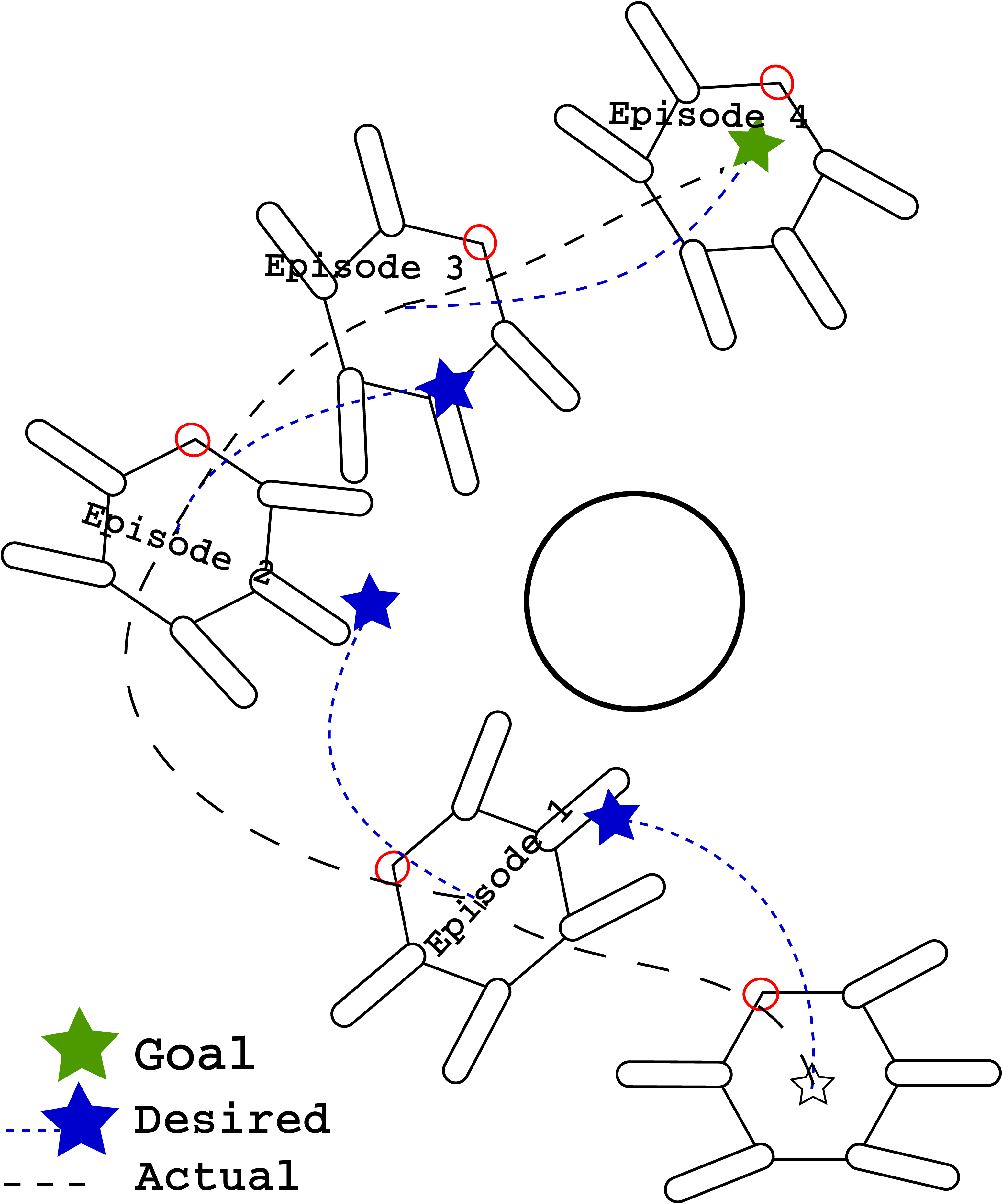}
    \caption{\label{fig:concept_exp}}
  \end{subfigure}


  \caption{Episodic vs Semi-episodic learning for robot damage recovery. (\subref{fig:concept_episodic}) \textbf{Episodic Learning}: The robot learns in episodes how to get to a single target starting from the same initial state. (\subref{fig:concept_exp}) \textbf{Semi-episodic Learning}: The robot learns the outcome of its atomic behaviors while executing the task.}

\end{figure}

Although the IT\&E approach is promising, its main limitation is the pure episodic approach it has adopted: for each trial (episode), the robot has to begin in the same starting state (Figure \ref{fig:concept_episodic}). This is limiting because learning a compensatory behavior has to be achieved in two steps, first learn a compensatory behavior, and then use it to complete the task. On the contrary, a wounded animal, for example, can perform trial and error ``episodes" to learn how to walk again, while going back to its nest for protection.

In this paper, we extend the IT\&E algorithm by (1) using a {\it generic reward} of the outcome of each atomic behavior of the robot in the adaptation part, and by (2) adding a {\it specialized reward selection layer} that selects a specialized reward function at each episode. These additions allow for a semi-episodic learning scheme that improves the robot's long-term autonomy by allowing to recover while attempting to achieve its task(s) (Figure \ref{fig:concept_exp}).

\section{Background}

\subsection{Bayesian Optimization with Gaussian Processes}

Bayesian Optimization (BO) is a well-established strategy for finding the extrema of functions that are expensive to evaluate \cite{brochu_tutorial_2010,mockus2013bayesian}. It is applicable in cases where one does not have a closed-form expression for the objective function (the function is a ``black-box''), but where one can obtain observations (possibly noisy) of this function. One of the distinctive features of BO is that it constructs a probabilistic model for the objective function and then exploits this model to make decisions about which point to evaluate next, while taking into account the uncertainty.

There are two major choices that must be made when performing BO. First, one must select a prior over functions that will express assumptions about the function being optimized. Second, one must choose an acquisition function, $u(\boldsymbol{x}|D_{1:t})$, which is used to construct a utility function from the model posterior, allowing us to determine the next point to evaluate.



Many models could be used for the BO prior, but Gaussian Process (GP) priors are the most common choice \cite{brochu_tutorial_2010}. A GP is an extension of the multivariate Gaussian distribution to an infinite-dimension stochastic process for which any finite combination of dimensions will be a Gaussian distribution \cite{brochu_tutorial_2010}. A GP is a distribution over functions, completely specified by its mean function, $m(\cdot)$ and covariance function, $k(\cdot,\cdot)$:

$$f(\boldsymbol{x}) \sim GP(m(\boldsymbol{x}), k(\boldsymbol{x}, \boldsymbol{x}'))$$

Assuming $D_{1:t} = \{(\boldsymbol{x_1}, f(\boldsymbol{x_1})),\cdots,(\boldsymbol{x_t}, f(\boldsymbol{x_t}))\}$ is a set of observations and $\sigma_{noise}^2$ the sampling noise, the GP is computed as follows:

$$p(f(\boldsymbol{x})|D_{1:t}, \boldsymbol{x}) = \mathcal{N}(m_t(\boldsymbol{x}), \sigma_t^2(\boldsymbol{x}))$$

where:
$$m_t(\boldsymbol{x}) = \boldsymbol{k^\top} \boldsymbol{K}^{-1}D_{1:t}$$
$$\sigma_t^2(\boldsymbol{x}) = k(\boldsymbol{x},\boldsymbol{x}) - \boldsymbol{k^\top} \boldsymbol{K}^{-1}\boldsymbol{k}$$
$$\boldsymbol{K} =
 \begin{bmatrix}
  k(\boldsymbol{x_1},\boldsymbol{x_1}) & \cdots & k(\boldsymbol{x_1},\boldsymbol{x_t}) \\
  \vdots  & \ddots & \vdots  \\
  k(\boldsymbol{x_t},\boldsymbol{x_1}) & \cdots & k(\boldsymbol{x_t},\boldsymbol{x_t})
 \end{bmatrix} + \sigma_{noise}^2I$$
$$\boldsymbol{k} = \begin{bmatrix} k(\boldsymbol{x},\boldsymbol{x_1}) & \cdots & k(\boldsymbol{x},\boldsymbol{x_t})\end{bmatrix}$$

We used {\it Upper Confidence Bound} (UCB) as the acquisition function. We refer the reader to {\it Brochu et al.} \cite{brochu_tutorial_2010} for a more detailed explanation.

\subsection{Intelligent Trial \& Error Algorithm}

IT\&E proposed a novel approach for robot damage recovery that consists of a 2-step process. An off-line evolutionary algorithm, \textbf{MAP-Elites} \cite{mouret_illuminating_2015}\cite{cully_robots_2015}, that generates many thousands of potential good behaviors is followed by a trial and error on-line adaptation part, based on BO (\textbf{M-BOA}), in order to find a compensatory behavior.

MAP-Elites is an evolutionary \emph{illumination} algorithm: instead of searching for a single, best solution, like optimization algorithms, MAP-Elites searches for the highest-performing individual for each point in a user-defined space. This user-defined space is often called the \textbf{behavior space}, because the dimensions of variation ({\it behavior descriptors}) usually measure behavioral characteristics. 


In IT\&E, the authors made a slight modification to the classical BO scheme. Their BO variation, called {\it Map-Based BO Algorithm} (M-BOA), models the difference between a {\it mean} function and the actual performance, instead of directly modeling the objective function ($\mathbf{P}(\cdot)$ is the {\it mean} function):

$$m_t(\boldsymbol{x}) = \mathbf{P}(\boldsymbol{x})+\boldsymbol{k^\top}\boldsymbol{K^{-1}}(D_{1:t}-\mathbf{P}(\boldsymbol{x_{1:t}}))$$

In the original work, the {\it mean} function was the prediction of the performance in the map generated from MAP-Elites. Algorithm \ref{mboa_algo} shows the pseudo-code for M-BOA.

\begin{algorithm}
\caption{M-BOA (Map-Based BO Algorithm)}\label{mboa_algo}
\begin{algorithmic}[1]
\Procedure{M-BOA}{}
\State $\forall \boldsymbol{x}\in map:$
\State\hspace{\algorithmicindent} $p(f(\boldsymbol{x})|\boldsymbol{x}) = \mathcal{N}(\mathbf{P}(\boldsymbol{x}), k(\boldsymbol{x},\boldsymbol{x}))$
\While {stopping criteria not met}
\State $\boldsymbol{x_{t+1}} = argmax_{\boldsymbol{x}} u(\boldsymbol{x}|D_{1:t})$ \Comment Next test point
\State $Y_{t+1} = performance(execute\_behavior(\boldsymbol{x_{t+1}}))$
\State $D_{1:t+1} = \{D_{1:t}, (\boldsymbol{x_{t+1}}, Y_{t+1})\}$
\State Update GP
\EndWhile
\EndProcedure
\end{algorithmic}
\end{algorithm}

\section{Approach}

\subsection{Generic Reward}

In the original IT\&E paper, the GP modeled the performance of each atomic behavior given a task. In this paper, we suggest learning a mapping from the atomic behaviors to the resulting relative outcomes. We call it a {\it Generic Reward} (GR) of the outcome of each atomic behavior of the robot. We use one GP for each dimension of the GR.

For example, imagine we have a robot moving in 2D space using an 1D continuous atomic behavior (direction to move $0.1$-step). A GR could be the relative position of the robot after executing a behavior - $(x,y)$. Thus, we need 2 GPs: $GP_x(\theta), GP_y(\theta)$. If we query the GPs at the point $\theta_0$, then we get a position, $p_1 = (GP_x(\theta_0), GP_y(\theta_0))$, as the prediction. In that way, we can now compute {\it specialized rewards} for different locomotion tasks, like the distance to different target points.

Put differently, the GR is a description of the outcome of each atomic behavior of the robot that it is generic-enough to be independent from one task to another, but specific-enough so that the performance of the outcome of one atomic behavior given a task can be computed.



The changes for {\it M-BOA} to work are:

\begin{itemize}
  \item define a {\it Reward} function that takes the GPs' prediction as input and returns the expected task performance;
  \item define an {\it Aggregator} function ({\it afun}) that takes as input the execution of an atomic behavior and returns the GR.
\end{itemize}

\subsection{Specialized Reward Selection Layer}

We, also, augment the proposed algorithm, by adding a layer responsible for selecting the {\it Reward} function, defined above. We call it {\it Specialized Reward Selection Layer} (RSL). Since we are modeling a GR of the outcome of each behavior of the robot and not the actual performance (given a task), we can change the {\it Reward} function as often as needed. This is true, because only the acquisition function needs an actual reward to select a new test point. We suggest updating or selecting the {\it Reward} function at each iteration of M-BOA.

For instance, if we consider the previous mobile robot example, at each iteration a planner algorithm chooses the next best point to reach. This point can then be used by the RSL in order to update the {\it Reward} function so that it outputs the Euclidean distance between the point selected by the planner and the prediction of the GPs.

\subsection{Semi-Episodic Learning Algorithm}

Using the two proposed additions, we can now have a non purely episodic version of the IT\&E algorithm. We call it {\bf Semi-Episodic Learning Algorithm} (SELA). The pseudo-code is shown in Algorithm \ref{ite_2_algo}.

\begin{algorithm}
\caption{Semi-Episodic Learning Algorithm}\label{ite_2_algo}
\begin{algorithmic}[1]
\Procedure{SELA}{}
\State {\bf Before mission} (in simulation with intact robot):
\State\hspace{\algorithmicindent} {\small Create Behavior-Performance Map using MAP-Elites}
\While {{\bf in mission}}
\If {significant performance drop}
\State Adaptation-Step (using {\bf SELA-ADAPT})
\EndIf
\EndWhile
\EndProcedure
\Procedure{SELA-ADAPT}{}
\State $\forall \boldsymbol{x}\in map:$
\State\hspace{\algorithmicindent} $p(f(\boldsymbol{x})|\boldsymbol{x}) = \mathcal{N}(\mathbf{P}(\boldsymbol{x}), k(\boldsymbol{x},\boldsymbol{x}))$
\While {stopping criteria not met}
\State Update {\it Reward} function
\State $\boldsymbol{x_{t+1}} = argmax_{\boldsymbol{x}} u(Reward(GPs(\boldsymbol{x}))|D_{1:t})$
\State $\boldsymbol{Y_{t+1}} = afun(execute\_behavior(\boldsymbol{x_{t+1}}))$
\State $D_{1:t+1} = \{D_{1:t}, (\boldsymbol{x_{t+1}}, \boldsymbol{Y_{t+1}})\}$
\State Update GPs
\EndWhile
\EndProcedure
\end{algorithmic}
\end{algorithm}

\section{Preliminary Experiments}

\subsection{Toy Simulation}

As a toy example, we consider the mobile robot example introduced previously. This mobile robot is a point (no dimensions, no orientation) and can take a 0.1-long step in any direction. We represent each atomic behavior by a scalar value, $\theta$: the direction of the corresponding move. This environment was inspired by {\it Engel et al.} \cite{engel_reinforcement_2005}. The task of the robot is to reach a target point despite some damage.

Because the example is too simple, but also to show the effectiveness of our method without relying on simulated data, we did not generate any behavior-performance map. We used the exact model of the intact robot as the mean function. Also, for the GR, we used the $(x,y)$ relative end position of each behavior, for the {\it Reward} function the Euclidean distance between the next target and the prediction of the GPs and for the reward selection layer an A* path planner.

To evaluate our technique we used the following two control experiments:

\begin{itemize}
  \item learn the model of the robot (using GPs) via random babbling and then use it to complete the task;
  \item solve the problem with the classic IT\&E approach: we first learn with IT\&E how to walk in 4 major directions (up, down, right, left) and then use these behaviors to reach the target.
\end{itemize}

\begin{figure}[!ht]
  \centering
  \includegraphics[scale=0.45]{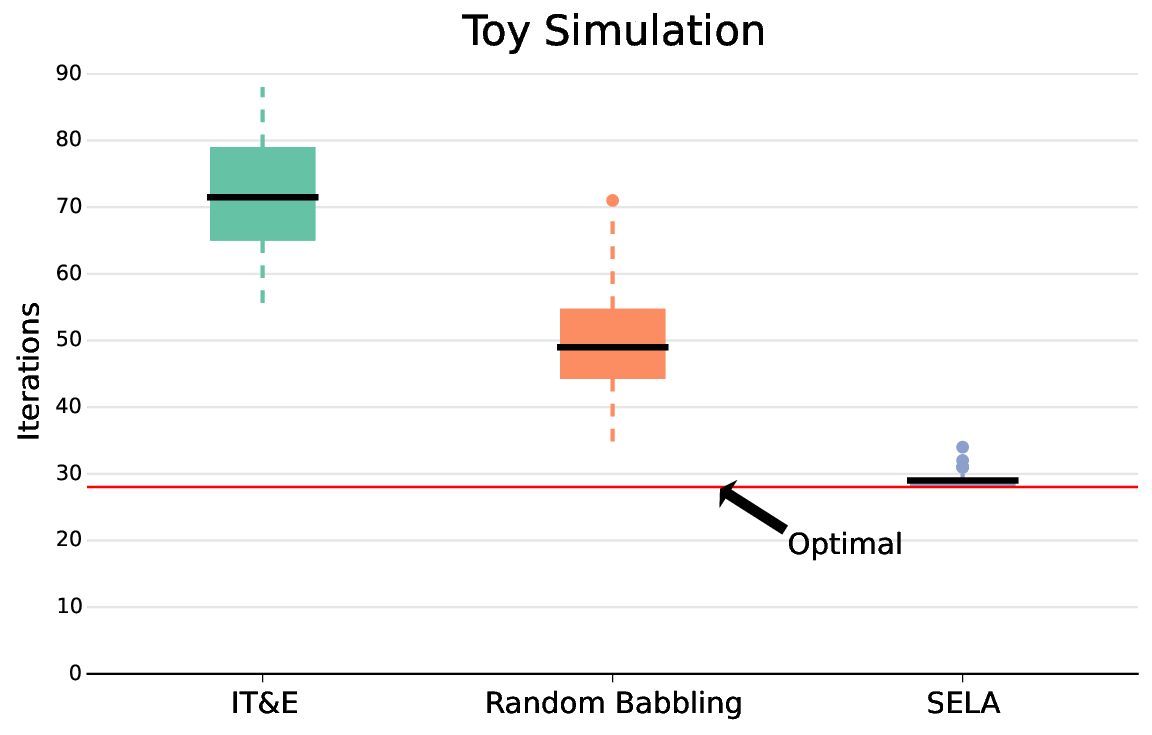}
  \caption{\textbf{Toy Simulation Evaluation}. Comparison between the baseline approaches and \textbf{SELA} for the toy simulation. For both of the baseline approaches, we measure the number of iterations required to learn and the number of steps that they take to complete the task. We ran 50 replicates of each approach.}
  \label{fig:toy_sim}
\end{figure}

We ran 50 replicates of each approach for the scenario: ``Reach the target point $(2.0,2.0)$ starting from the origin despite a $0.5$ radians angle offset in the range direction $\theta > 0$". To make the task a little more realistic we added a small Gaussian noise ($\mu = 0, \sigma^2 = 0.01$) to the position observations. Figure \ref{fig:toy_sim} shows the resulting performance (number of atomic behaviors taken to reach the target) for the different approaches. Our algorithm is able to reach the target with almost the optimal number of steps (i.e. if we perfectly knew the model), that is in much fewer steps than the other approaches.

\subsection{6-Legged Simulated Robot locomotion task}

As a more realistic example, we consider a simulated 6-legged (hexapod) robot moving in space with the same task as in the toy simulation. See Figure \ref{fig:concept_exp} for the scenario and  \cite{cully_robots_2015} for more details on the simulated hexapod. We evolved different atomic behaviors using the MAP-Elites algorithm with an 8D behavior descriptor (2 dimensions for space diversity + 6 dimensions for walking diversity), inspired by \cite{cully_behavioral_2013,cully2015creative}. The number of atomic behaviors evolved were approximately {\it 1 million}. We used this behavior-performance map as the mean function. All the other parameters were the same as in the toy simulation experiment.


To evaluate our technique we used similar control experiments as in the toy simulation experiment:

\begin{itemize}
  \item IT\&E variant \#1: we learn the outcome of the atomic behaviors (using GPs) via selecting the most uncertain behavior for $N=15$ iterations. This can be considered as a uniform sampling of the behavior space. We then use what we learned to reach the target.
  \item IT\&E variant \#2: we first learn with IT\&E how to walk in 4 major directions (forward, backward, turn cw, turn ccw) and then use these behaviors to reach the target.
\end{itemize}

\begin{figure}[!ht]
  \centering
  \includegraphics[scale=0.45]{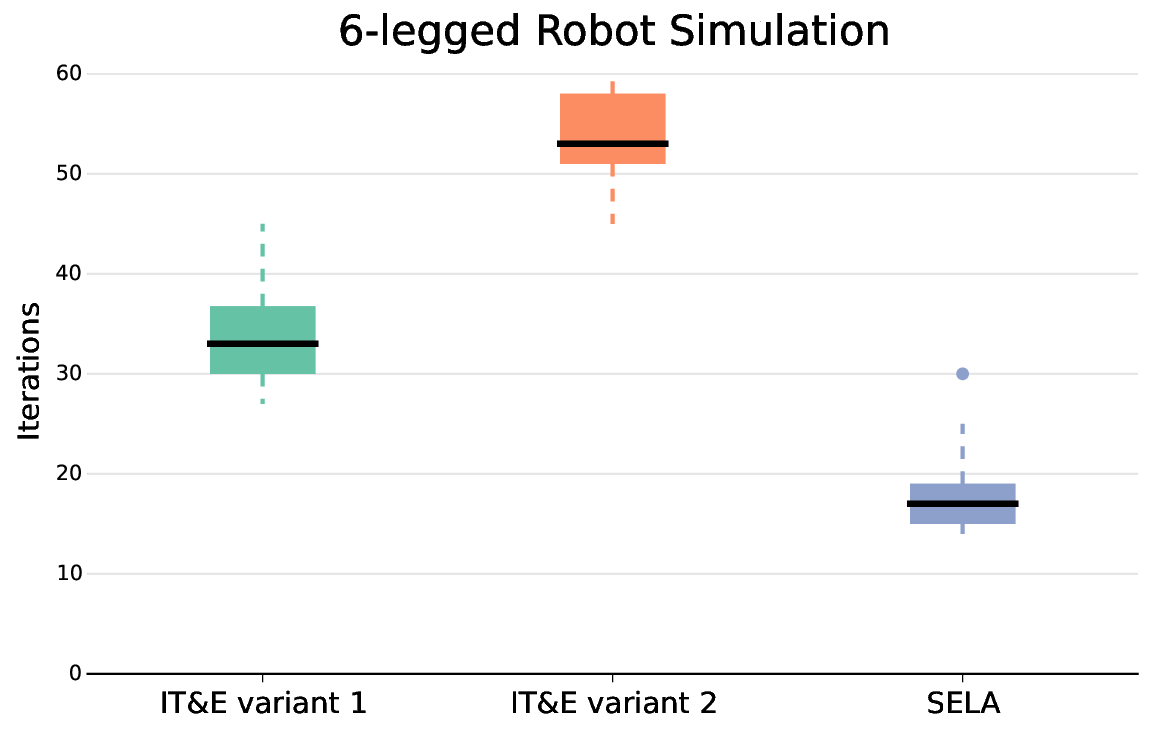}
  \caption{\textbf{6-legged Robot Simulation Evaluation}: Comparison between the baseline approaches and \textbf{SELA} for the 6-legged robot simulation. For both of the baseline approaches, we measure the number of the iterations required to learn and the number of steps that they take to complete the task. We ran 50 replicates of each approach.}
  \label{fig:hexapod_eval}
\end{figure}

We ran 50 replicates of each approach for the scenario: ``Reach the target point $(2.0,2.0)$ despite the middle right leg being removed". We, also, added a small Gaussian noise ($\mu = 0, \sigma^2 = 0.01$) to the position observations. Figure \ref{fig:hexapod_eval} shows the resulting performance (number of atomic behaviors taken to reach the target) for the different approaches. Our algorithm is able to find solutions in fewer steps than the other approaches.

\subsection{6-Legged Robot locomotion task}

We, also, applied our technique on a real 6-legged robot. Preliminary experiments show promising results\footnote{{\small\url{https://www.youtube.com/watch?v=Gpf5h07pJFA}}}.

\section{Conclusion and Future Work}

We have introduced a semi-episodic learning scheme for robot damage recovery and a novel algorithm in this direction: \textbf{Semi-Episodic Learning Algorithm}. The intuition behind this scheme is that the robot can learn in a {\it data-efficient} way how to compensate for damages \emph{while completing its task(s)}. This is achieved by (1) shrinking the search space, using simulated or computed data as prior knowledge, and by (2) using a {\it generic reward} of the outcome of the atomic behaviors of the robot instead of their performance given a task.

Future work includes performing more experiments with the real robot as well as experiments with different robots. In addition, BO can be replaced by other techniques that scale better. What is more, we used a naive {\it reward selection layer}, but more efficient/sophisticated methods can be used. We are currently investigating in this direction. Additionally, theoretical guarantees and analysis should be investigated in detail. Overall, this work is a first step towards semi-episodic and life-long learning for robot damage recovery.


\addtolength{\textheight}{-12cm}   



\section*{Appendix}
\begin{footnotesize}
For all experiments the following parameters were used:

\textbf{Error threshold for reaching goal:} $\epsilon_{goal} = 0.1$

\subsection{BO with GPs}

\textbf{Acquistion function:} UCB with $\alpha = 0.05$

\textbf{Kernel:} Exponential kernel with $\sigma = 0.1$

\textbf{GP noise:} $\sigma_{noise}^2 = 0.001$

\textbf{Max iterations:} $N = 10$ (Toy), $N = 15$ (Hexapod)

\subsection{Learning with random babbling}

\textbf{Error threshold:} $\epsilon_{model} = 0.01$

\textbf{Max iterations:} $N = 15$

%
%
%
%
%


\end{footnotesize}

\section*{Acknowledgments}

This work was supported by the \textbf{ERC project ``ResiBots"} ({\it grant agreement No 637972}), funded by the {\it European Research Council}.

%
%
%
%

\bibliographystyle{IEEEtran}
\bibliography{IEEEabrv,mybib}

\end{document}